\documentclass[runningheads]{llncs}
\usepackage{amsmath}
\usepackage[T1]{fontenc}
\usepackage{amsfonts}
\usepackage{cleveref}
\usepackage{balance}
\usepackage[decisionutilitycolor]{influence-diagrams2}
\usepackage{float}
\usepackage{microtype}
\usepackage[normalem]{ulem}
\useunder{\uline}{\ul}{}
\usepackage{caption}
\usepackage{subcaption}
\usepackage{graphicx}
\usepackage{bm}

\usepackage[disable]{todonotes}

\begin{document}

\title{Characterising Decision Theories with Mechanised Causal Graphs}

 \author{Matt MacDermott\inst{1}\and
 Tom Everitt\inst{2} \and
 Francesco Belardinelli\inst{1}}

 \authorrunning{M. MacDermott et al.}

 \institute{
 Imperial College London \email{m.macdermott21@imperial.ac.uk} \and
 Google DeepMind \email{tomeveritt@deepmind.com} \and
 Imperial College London \email{francesco.belardinelli@imperial.ac.uk}
\\
}

\maketitle              %
\begin{abstract}

How should my own decisions affect my beliefs about the outcomes I expect to achieve? If taking a certain action makes me view myself as a certain type of person, it might affect how I think others view me, and how I view others who are similar to me. This can influence my expected utility calculations and change which action I perceive to be best. Whether and how it should is subject to debate, with contenders for how to think about it including evidential decision theory, causal decision theory, and functional decision theory.
In this paper, we show that mechanised causal models can be used to characterise and differentiate the most important decision theories, and generate a taxonomy of different decision theories.

\keywords{Decision Theory \and Causality \and Game Theory.}
\end{abstract}

\section{Introduction}

What's the best course of action, given a particular objective, and some beliefs about the present situation?
To make this question more concrete, decision theory typically represents beliefs with a probability distribution or causal model, and objectives with a real-valued utility function, such that the preferred course of action is to maximise the expected utility.
Even under these constraints, the "correct" decision procedure remains subject to debate.
Should expected utility be calculated by conditioning on the action \cite{jeffrey_logic_1990,ahmed_evidence_2014}, intervening on the action \cite{lewis_causal_1981,skyrms_causal_1982,gibbard_counterfactuals_1981,joyce_foundations_1999}, intervening on the policy \cite{yudkowsky_functional_2018,meacham_binding_2010}, or by some other method \cite{dai_towards_2009,soares_toward_2015}? And in what kind of model?

The choice of procedure matters especially in situations where my very action can teach me something about myself. On the one hand, learning about myself informs me about the behaviour of similar agents -- so that if I find myself in a prisoner's dilemma with one, for example, my decision about whether to co-operate tells me something about theirs. My decision also informs me about what others might think of me -- so that it helps me know what to expect from anybody who is making sufficiently accurate predictions about me.

The part of the decision theory debate we are interested in is how and when it is appropriate to let these sort of inferences influence one's decision. This question is particularly relevant for AI systems. Whilst we humans rarely encounter exact copies of ourselves or oracular predictors of our behaviour, AI systems have source code (or model weights) that can be duplicated or inspected by other agents. We might therefore expect them to frequently find themselves in the sort of situations described in decision-theoretic thought experiments. 

This has several important consequences. Firstly, if we are going to be delegating our decision-making to artificial agents, we had better be clear on what constitutes good decision-making. Understanding the pros and cons of different decision theories can help us design agents that handle tricky situations adequately.

Secondly, powerful AI systems may be dangerous \cite{Bostrom2014,carlsmith2022power}. Understanding how a system's design and training influences its decision theory \cite{krueger2020hidden,bell_reinforcement_2021} may help us anticipate and avoid situations where things go wrong. To illustrate this point, note that several proposals for aligning AI systems involve using multiple agents to police each other's behaviour, in the hopes that a Nash equilibrium of mutual defection emerges to keep the various parts of the system honest \cite{irving_ai_2018,leike_scalable_2018}. But as we will see in the case of the prisoner's dilemma, some decision theories can be used to \emph{avoid} Nash equilibria to the mutual benefit of the agents involved. An understanding of decision theory can help us avoid potentially catastrophic situations where powerful agents meant to keep each other in check collude against their overseers.

\paragraph{Contributions}
In this paper we show that mechanised causal models \cite{hammond_reasoning_2023,kenton_discovering_2022} offer a unified framework for comparing and analysing different decision theories.
In particular, standard decision theory problems can be usefully modelled using mechanised causal graphs (Section \ref{modelling}), and common decision theories can be characterised and distinguished based on which nodes they condition and intervene on (Section \ref{theories}).
This in turn allows us to identify dimensions along which decision theories differ, and propose a taxonomy of available decision principles (Section \ref{taxonomy}). Finally, we use mechanised graphs to clarify when a certain class of decision problem is well-defined (Section \ref{defining}). 

\section{Modeling Decision Problems}
\label{modelling}

In this section, we describe some decision problems often used to distinguish different principles for decision-making, and how to model them with causal graphs.

\subsection{Decision Problems}

Several thought experiments have long been used in the decision theory literature to probe at intuitions around whether a rational agent should choose actions based on the evidence they provide about expected outcomes, or based on their causal effects.
We start by considering the following Newcomb's Problem, as it first appeared in \cite{nozick_newcombs_1969}.

\begin{example}[Newcomb's Problem]
\label{example:newcomb}
You stand before two boxes. One is transparent and contains one thousand dollars; the other is opaque and either contains either one million dollars or nothing. Your choice is between taking both boxes ("two-boxing") and taking just the opaque box ("one-boxing"). A reliable predictor, who has been right in $99\%$ of previous cases, has put the million in the opaque box if and only if it predicted you would one-box.
\end{example}

Newcomb's problem gives its name to the class of `Newcomblike' problems. Such problems arise when your action provides evidence about the state of the world beyond its causal effect. A variant of Newcomb's problem makes both boxes transparent \cite{gibbard_counterfactuals_1981}, which will help to illustrate considerations about how you should use your observations to update your beliefs:

\begin{example}[Transparent Newcomb]
    You face the same choice as in Newcomb's problem, except that now both boxes are transparent, and the predictor has put the million in the second box if and only if they predicted you would one-box upon seeing it.
\end{example}

Finally, we introduce the a version of the Twin Prisoner's Dilemma \cite{lewis1979prisoners}, to illustrate subtleties in the definition of causality:

\begin{example}[Twin Prisoner's Dilemma]
    You face a Prisoner's Dilemma. You and your opponent will each co-operate or defect. You rank the outcomes from best to worse as follows: defecting against co-operation, mutual co-operation, mutual defection, being defected against while co-operating. You can see that your opponent is an agent with identical source code to your own. You know that the source code will compile into the same policy for you as for your opponent, but the only way to determine what the policy actually does is to decide your action.
\end{example}

\subsection{Causal Models}

The different decision problems can be modeled using Pearl's causal models \cite{pearl_causality_2000}, which consists of three kinds of probabilistic models: \textit{associational}, \textit{interventional}, and \textit{counterfactual}.

We will use capital letters ($V$) for variables, lowercase letters ($v$) for instantiations of variables, and bold letters ($\bm{V}$ or $\bm{v}$) for sets of the above. We write $dom(V)$ for the set of oucomes of a random variable $V$.

\begin{definition}[Bayesian Network]
A \emph{Bayesian network} over a set of random variables $\bm{V}$ with joint distribution $Pr(\bm{V})$ is a directed acyclic graph (DAG) $\mathcal{G}=(\bm{V},\mathcal{E})$ with vertices $\bm{V}$ and edges $\mathcal{E}$ such that the joint distribution can be factorised as $Pr(\bm{V})=\prod_{V\in\bm{v}}Pr(V\mid \bm{Pa_V})$, where $\bm{Pa_V}$ are the parents of $V$ in $G$.
\end{definition}

\begin{figure}
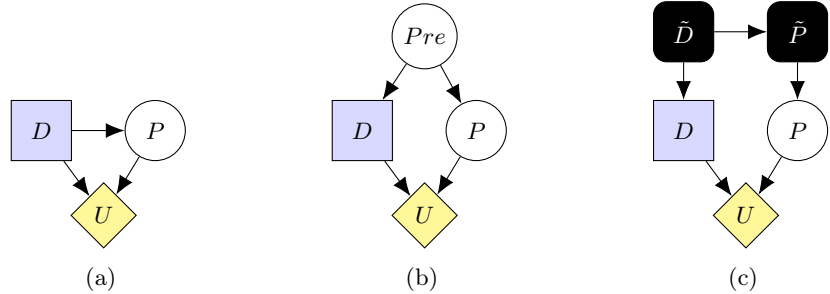

     \centering
     \begin{subfigure}[b]{0.3\linewidth}
         \centering
         \begin{influence-diagram}
         
    \node (D) [decision] {$D$};
    \node (P) [right = of D] {$P$};
    \node (U) [below right = 0.5 and 0.2 of D, utility] {$U$};

    \edge {D} {U};
    \edge {D} {P};
    \edge {P} {U};
    \end{influence-diagram}
         \caption{}
         \label{fig:Newcomb-EDT}
     \end{subfigure}
     \hfill
     \begin{subfigure}[b]{0.3\linewidth}
         \centering
    \begin{influence-diagram}
    \node (D) [decision] {$D$};
    \node (P) [right = of D] {$P$};
    \node (U) [below right = 0.5 and 0.2 of D, utility] {$U$};
    \node (Pre) [above =1.5 of U] {$Pre$};

    \edge {D} {U};
    \edge {Pre} {D};
    \edge {Pre} {P};
    \edge {P} {U};
    \end{influence-diagram}
         \caption{}
         \label{fig:Newcomb-CDT}
     \end{subfigure}
     \hfill
     \begin{subfigure}[b]{0.3\linewidth}
         \centering
         \begin{influence-diagram}
    \node (D) [decision] {$D$};
    \node (P) [right = of D] {$P$};
    \node (U) [below right = 0.5 and 0.2 of D, utility] {$U$};
    \node (Dt) [relevanceb, above = 0.5 of D] {$\tilde{D}$};
    \node (Pt) [relevanceb, above = 0.5 of P] {$\tilde{P}$};

    \edge {D} {U};
    \edge {Dt} {D};
    \edge {Pt} {P};
     \edge {Dt} {Pt};
    \edge {P} {U};
    \end{influence-diagram}
         \caption{}
         \label{fig:Newcomb-mech}
     \end{subfigure}

        \caption{Newcomb's problem as (a) Bayesian network, (b) a causal Bayesian network, and (c) a mechanised causal Bayesian network with $\tilde U$ omitted.}
        \label{fig:Newcomb}
\end{figure}

We borrow from the literature on influence diagrams \cite{howard2005influence,everitt_agent_2021} the convention of using square nodes for decision variables, diamond nodes for utility variables, and round nodes for everything else. However, it is important to remember that our models are \textit{not} influence diagrams, since we assume probability distributions over decision variables rather than leaving them blank. This introduces the question of how such probability distributions should feature in expected utility calculations, which is precisely the debate about causal and evidential decision theory.

Newcomb's problem can be modeled with a Bayesian network as in ~\cref{fig:Newcomb-EDT}, where we have variables representing the agent's decision $D$, the prediction $P$, and the utility $U$. The association between the agent's decision and the prediction is represented with the (non-causal) edge $D\to P$.
Note that while this edge doesn't represent a causal relationship, nor would a reverse edge $P\to D$, as the prediction does not causally influence the decision (changing the prediction would not change the decision).
The utility $U$ depends on both the decision $D$ and the prediction $P$, which is modelled with the (causal) edges $D\to U\gets P$.

Causal Bayesian networks \cite{pearl_causality_2000} are a particular kind of Bayesian network that can represent the effects of external interventions, effectively by requiring that edges always represent direct causal influences. An intervention on a set of variables can then be understood formally as replacing the conditional probability distributions of those variables, which changes the marginal distributions of causally downstream variables without affecting those upstream.

\begin{definition}[Causal Bayesian Network]
A \emph{causal Baysian network} over a set of random variables $\bm{V}$ with joint interventional distributions $\{Pr(\bm{V}\mid do(\bm{Y}=\bm{y})\}$ is a Bayesian network $\mathcal{G}=(\bm{V},\mathcal{E})$ such that for any $\bm{Y}\subseteq\bm{V}$ and $\bm{y}\in dom(\bm{Y})$, the joint distribution equals
$$Pr(\bm{V}\mid do(\bm{Y}=\bm{y})) = \prod_{V\in\bm{V}\setminus\bm{Y}}Pr(V\mid \bm{Pa}_V)$$
for values of $\bm{V}$ compatible with $\bm{y}$, and values of $\bm{V}$ incompatible with $\bm{y}$ have probability $0$.
\end{definition}

\Cref{fig:Newcomb-CDT} shows Newcomb's problem represented as a Causal Bayesian network.
To make all arrows represent causal relationships, it is necessary to introduce a confounding variable representing the predisposition of the agent to one-box or two-box.
The other variables are as in \cref{fig:Newcomb-EDT}.

A more systematic way of representing agents' predispositions is to include nodes for their policies, as well as for the mechanisms governing other variables in the graph \cite{kenton_discovering_2022,hammond_reasoning_2023}.
This can be done with a special kind of causal Bayesian network called a \textit{mechanised} causal Bayesian network.

\begin{definition}[Mechanised Causal Bayesian Network]
A \emph{mechanised causal Bayesian network} is a causal Bayesian network over a set of variables $\bm{\mathcal{V}}$ which is partioned into \emph{object-level variables} $\bm{V}$ and \emph{mechanism-level variables} $\bm{\tilde{V}}$. Each object-level variable $V\in\bm{V}$ has a single mechanism parent $\tilde{V}\in\bm{\tilde{V}}$, such that the value of $\tilde{V}$ sets the probability distribution $Pr(V\mid \bm{Pa_V})$, where $\bm{Pa_V}$ is the set of object-level parents of $V$. 
\end{definition}

We call the mechanism of a decision variable a \emph{decision rule} variable. An agent's policy is a tuple consisting of the decision rules for each of its decision nodes. In this paper we restrict our attention to settings with a single decision variable, and so we will use the terms somewhat interchangeably \footnote{Following the convention of the decision theory literature, we usually restrict agents to deterministic policies. Stochastic policies can be modelled by allowing the agent to observe a source of randomness before making a decision.}

Newcomb's problem is represented with a mechanised causal graph in \cref{fig:Newcomb-mech}, obtained from \cref{fig:Newcomb-CDT} by adding mechanism parents $\tilde{D},\tilde{U}$, and $\tilde{P}$ for each of the object-level variables $D,U,$ and $P$ (since the utility mechanism $\tilde{U}$ is not relevant to our analysis, we suppress it in our diagrams). We will no longer need the confounding "predisposition" variable to explain the correlation of the agent's decision with the prediction. Instead, we can model the connection between the prediction and the agent's behaviour by letting the mechanism of the prediction variable depend on the mechanism variable of the of the agent's decision.

\begin{figure}
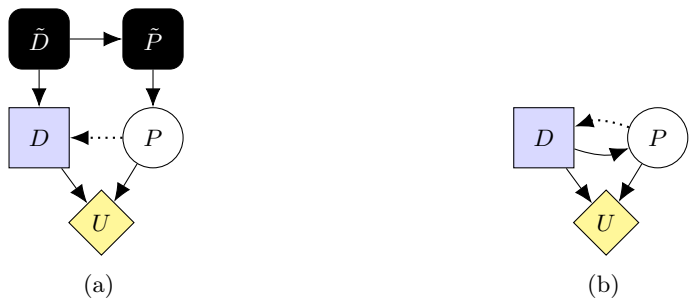

     \centering
 
     \begin{subfigure}[b]{0.45\linewidth}
         \centering
          \begin{influence-diagram}
    \node (D) [decision] {$D$};
    \node (P) [right = of D] {$P$};
    \node (U) [below right = 0.5 and 0.2 of D, utility] {$U$};
    \node (Dt) [relevanceb, above = 0.5 of D] {$\tilde{D}$};
    \node (Pt) [relevanceb, above = 0.5 of P] {$\tilde{P}$};

    \edge {D} {U};
    \edge {Dt} {D};
    \edge {Pt} {P};
     \edge {Dt} {Pt};
    \edge {P} {U};
    \edge[information] {P} {D};
    \end{influence-diagram}
         \caption{}
         \label{fig: mech TN}
     \end{subfigure}
     \hfill
     \begin{subfigure}[b]{0.45\linewidth}
         \centering
           \begin{influence-diagram}
    \node (D) [decision] {$D$};
    \node (P) [right = of D] {$P$};
    \node (U) [below right = 0.5 and 0.2 of D, utility] {$U$};

    \edge {D} {U};
    \path (D) edge[->, bend right=20] (P);
    \edge {P} {U};
    \path (P) edge[information, ->, bend right=20] (D);
    \end{influence-diagram}
         \caption{}
         \label{fig: cyclic TN}
     \end{subfigure}
        \caption{(a) Transparent Newcomb, and (b) Transparent Newcomb where the predictor predicts the decision rather than the decision rule.}
        \label{fig:Transparent Newcomb}
\end{figure}

The Transparent Newcomb problem can be modeled in much the same way, but with an edge added from $P$ to $D$, representing the fact that the agent observes the prediction, which means that the prediction is now able to causally influence the decision (see \cref{fig:Transparent Newcomb}). As in the causal influence diagram literature, we will used dashed edges to represent agents' observations.

\subsection{Notions of Causality}
\label{sec:notions-of-causality}

Finally, the Twin Prisoner's Dilemma brings up an important distinction about causality.
Abstractly, a causal model over a set of variables needs to specify two operations for each variable: how to intervene, and how to measure.
Once these operations are specified, a well-defined probability distribution is generated for each possible combination of interventions.

\begin{figure}
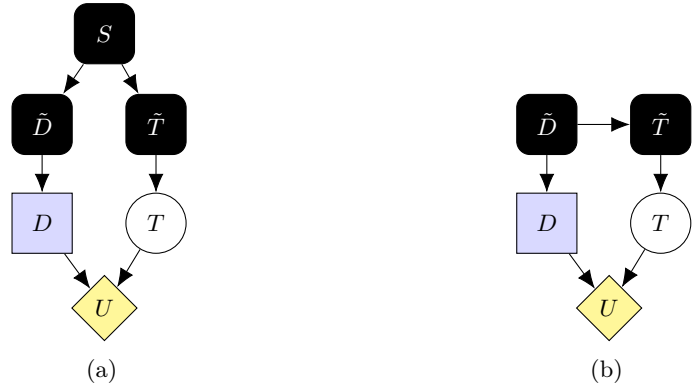

     \centering
 
     \begin{subfigure}[b]{0.45\linewidth}
         \centering
             \begin{influence-diagram}
    \node (D) [decision] {$D$};
    \node (T) [right = of D] {$T$};
    \node (U) [below right = 0.5 and 0.2 of D, utility] {$U$};
    \node (Dt) [relevanceb, above = 0.5 of D] {$\tilde{D}$};
    \node (Tt) [relevanceb, above = 0.5 of T] {$\tilde{T}$};
    \node (S) [relevanceb, above = 2.8  of U] {$S$};

    \edge {D} {U};
    \edge {Dt} {D};
    \edge {Tt} {T};
     \edge {S} {Dt};
     \edge {S} {Tt};
    \edge {T} {U};
    \end{influence-diagram}
         \caption{}
         \label{fig:TwinPD CDT}
     \end{subfigure}
     \hfill
     \begin{subfigure}[b]{0.45\linewidth}
         \centering
         
\begin{influence-diagram}
    \node (D) [decision] {$D$};
    \node (T) [right = of D] {$T$};
    \node (U) [below right = 0.5 and 0.2 of D, utility] {$U$};
    \node (Dt) [relevanceb, above = 0.5 of D] {$\tilde{D}$};
    \node (Tt) [relevanceb, above = 0.5 of T] {$\tilde{T}$};

    \edge {D} {U};
    \edge {Dt} {D};
    \edge {Tt} {T};
     \edge {Dt} {Tt};
    \edge {T} {U};
    \end{influence-diagram}
    \caption{}
    \label{fig:TwinPD FDT}
    \end{subfigure}
        \caption{The Twin Prisoner's Dilemma (a) from the perspective of a causal decision theorist and (b) from the perspective of a functional decision theorist.}
        \label{fig:Twin PD}
\end{figure}

In the study of decision problems, the operations we have in mind are often sufficiently obvious from an informal description of the variables that we do not need to explicitly specify them. 
Indeed, this is the case for the prediction, decision, and utility in \cref{fig:Newcomb,fig:Transparent Newcomb}.
But in the Twin Prisoner's Dilemma, one might interpret the policy node in two different ways, and the interpretation will affect the causal structure.
We could interpret intervening on your policy $\tilde D$ as changing the physical result of the compilation of your source code, such that an intervention will only affect your decision $D$, and not that of your twin $T$. Under this physical notion of causality, we get \cref{fig:TwinPD CDT}, where there is a common cause $S$ explaining the correlation between the agent's policy and its twin's.

But on the other hand, if we think of intervening on your policy as changing the way your source code compiles in all cases, then intervening on it will affect your opponent's policy, which is compiled from the same code.
In this case, we get the structure shown in \cref{fig:TwinPD FDT}, where an intervention on my policy would affect my twin's policy\footnote{We could alternatively imagine mutual causation between the agent's policy and the twin's policy, but for the purposes of this paper we will model the world from the subjective first-person perspective of a particular agent.}.
We can view this as an intervention on an abstract "logical" variable rather than an ordinary physical variable. We therefore call the resulting model a \emph{logical-causal} model.

Pearl's notion of causality is the physical one, but Pearl-style graphs have also been used in the decision theory literature to represent logical causality \cite{soares_toward_2015,yudkowsky_functional_2018,levinstein_cheating_2020}. One purpose of this paper
is to show that mechanism variables are a useful addition to \textit{any} graphical model being used in decision theory.

\section{Decision Theories Using Mechanised Causal Graphs}

\label{theories}

In this section we use mechanised causal graphs to describe three competing approaches to calculating the expected utility of a course of action: causal decision theory (CDT), evidential decision theory (EDT), and functional decision theory (FDT). While causal graphs have often been used to characterise the difference between causal and evidential expected utility calculations, the introduction of mechanism variables allows us to go a step further and define FDT's feature of "updatelessness" as a property of grapical queries.

\subsection{Causal Decision Theory}

Causal decision theory \cite{joyce_foundations_1999,gibbard_counterfactuals_1981,skyrms_causal_1982,lewis_causal_1981} asks you to consider the causal effect of each of the options you have at your disposal.
CDT has various notable formalisations, and many arguments have been made in its favour, but perhaps its greatest appeal is that choosing an action for any reason other than its causal effect seems intuitively bizarre. 

\begin{quote}
\textbf{Slogan}: Choose the action with the best causal effect.
\end{quote}

In Newcomb's problem, if you only care about the causal effects of your actions, you should two-box. You will get the opaque box either way and its contents have been decided already; you cannot causally influence whether you get the million. You \textit{can} affect whether you get the thousand -- so your expected wealth is one thousand dollars higher under two-boxing than one-boxing. Things are similar in Transparent Newcomb and the Twin Prisoner's Dilemma -- your decision does not causally affect the contents of the boxes or whether your twin co-operates, so you should one-box and defect respectively.

\paragraph{Expected Utility Calculations} A causal decision theorist computes the expected utility of taking each action $d\in dom(D)$ as $$\mathbb{E}\left[ U\mid do(D=d),\bm{Obs}_D=\bm{obs}_D \right].$$

\begin{figure}
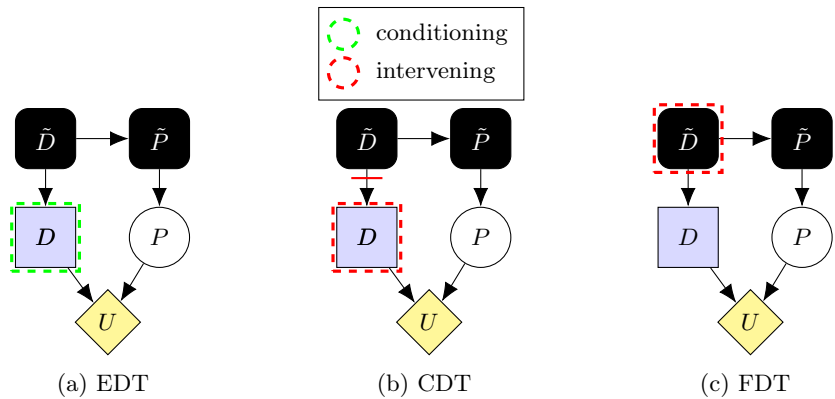

     \centering
     \begin{influence-diagram}
         \cidlegend{
 \legendrow{response incentive}{conditioning}\\
 \legendrow{feasible control incentive}{intervening} \\}
     \end{influence-diagram}
     
     \begin{subfigure}[b]{0.3\linewidth}
         \centering
             \begin{influence-diagram}
    \node (D) [decision] {$D$};
    \node (P) [right = of D] {$P$};
    \node (U) [below right = 0.5 and 0.2 of D, utility] {$U$};
    \node (Dt) [relevanceb, above = 0.5 of D] {$\tilde{D}$};
    \node (Pt) [relevanceb, above = 0.5 of P] {$\tilde{P}$};

    \edge {D} {U};
    \edge {Dt} {D};
    \edge {Pt} {P};
     \edge {Dt} {Pt};
    \edge {P} {U};

    \ri[rectangle]{D}
    \end{influence-diagram}
         \caption{EDT}
         \label{fig: Newcomb op EDT}
     \end{subfigure}
     \hfill
     \begin{subfigure}[b]{0.3\linewidth}
         \centering
        \begin{influence-diagram}
    \node (D) [decision] {$D$};
    \node (P) [right = of D] {$P$};
    \node (U) [below right = 0.5 and 0.2 of D, utility] {$U$};
    \node (Dt) [relevanceb, above = 0.5 of D] {$\tilde{D}$};
    \node (Pt) [relevanceb, above = 0.5 of P] {$\tilde{P}$};

    \edge {D} {U};
    \edge {Dt} {D};
    \edge {Pt} {P};
     \edge {Dt} {Pt};
    \edge {P} {U};

    \node (crossout) at ($(Dt)!0.4!(D)$) [draw=none, minimum size=4mm] {};
    \draw[red, thick] (crossout.west) edge (crossout.east);

    \fci[rectangle]{D}
    \end{influence-diagram}
         \caption{CDT}
         \label{fig: Newcomb op CDT}
     \end{subfigure}
     \hfill
     \begin{subfigure}[b]{0.3\linewidth}
         \centering
         \begin{influence-diagram}
    \node (D) [decision] {$D$};
    \node (P) [right = of D] {$P$};
    \node (U) [below right = 0.5 and 0.2 of D, utility] {$U$};
    \node (Dt) [relevanceb, above = 0.5 of D] {$\tilde{D}$};
    \node (Pt) [relevanceb, above = 0.5 of P] {$\tilde{P}$};

    \edge {D} {U};
    \edge {Dt} {D};
    \edge {Pt} {P};
     \edge {Dt} {Pt};
    \edge {P} {U};
    \node at (Dt) [draw, rectangle, incentive, fci-color, minimum size=9mm] {};
    \end{influence-diagram}
         \caption{FDT}
         \label{fig: Newcomb op FDT}
     \end{subfigure}

        \caption{The operation performed by (a) an evidential decision theorist, (b) a causal decision theorist, and (c) a functional decision theorist on a graphical model of Newcomb's Problem.}
        \label{fig:Newcomb-decision-theories}
        
\end{figure}

\Cref{fig: Newcomb op CDT} shows the operation performed by CDT in a mechanised causal graph for Newcomb's problem. The do-intervention severs the edge from $\tilde{D}$ to $D$, so the value of $D$ tells us nothing about $P$, and the agent two-boxes. In Transparent Newcomb (\cref{fig:Transparent Newcomb}) and the Twin Prisoner's Dilemma (\cref{fig:TwinPD CDT}) the story is much the same.

\paragraph{Model Requirements.} CDT agents compute their expected utility under an intervention, so they require a model at the second rung of Pearl's causal hierarchy. And since no mechanism variables appear in the expression, the model does not need to be mechanised. But as we just saw, CDT agents can happily use models with mechanism variables. They can also use models at the third rung of the hierarchy.

In order to obtain a model of a decision problem sufficient to apply CDT, we need to estimate the probability distribution $P(U\mid do(D=d),\bm{Obs}_D)$ for each $d\in dom(D)$. In principle we can do this by performing a randomised controlled trial experiment: find sufficiently similar agents to ourselves facing the same decision problem, assign them an action to take at random, and record their observations and the utility they recieve. But we need to be able to make these assignments as an outsider to the system, so that for example, a Newcomblike predictor could not have foreseen our intervention and predicted the agent's action accordingly.

\begin{figure}
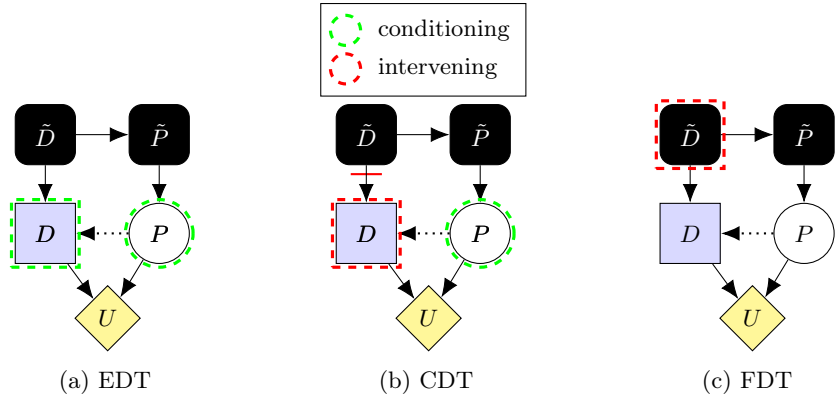

     \centering

    \begin{influence-diagram}
         \cidlegend{
 \legendrow{response incentive}{conditioning}\\
 \legendrow{feasible control incentive}{intervening} \\}
     \end{influence-diagram}
     
     \begin{subfigure}[b]{0.3\linewidth}
         \centering
             \begin{influence-diagram}
    \node (D) [decision] {$D$};
    \node (P) [right = of D] {$P$};
    \node (U) [below right = 0.5 and 0.2 of D, utility] {$U$};
    \node (Dt) [relevanceb, above = 0.5 of D] {$\tilde{D}$};
    \node (Pt) [relevanceb, above = 0.5 of P] {$\tilde{P}$};

    \edge {D} {U};
    \edge {Dt} {D};
    \edge {Pt} {P};
     \edge {Dt} {Pt};
    \edge {P} {U};
     \edge[information] {P} {D};

    \ri{P}
    \ri[rectangle]{D}
    \end{influence-diagram}
         \caption{EDT}
         \label{fig:Newcomb-mech EDT}
     \end{subfigure}
     \hfill
     \begin{subfigure}[b]{0.3\linewidth}
         \centering
        \begin{influence-diagram}
    \node (D) [decision] {$D$};
    \node (P) [right = of D] {$P$};
    \node (U) [below right = 0.5 and 0.2 of D, utility] {$U$};
    \node (Dt) [relevanceb, above = 0.5 of D] {$\tilde{D}$};
    \node (Pt) [relevanceb, above = 0.5 of P] {$\tilde{P}$};

    \node (crossout) at ($(Dt)!0.4!(D)$) [draw=none, minimum size=4mm] {};
    \draw[red, thick] (crossout.west) edge (crossout.east);

    \edge {D} {U};
    \edge {Dt} {D};
    \edge {Pt} {P};
     \edge {Dt} {Pt};
    \edge {P} {U};
     \edge[information] {P} {D};

    \ri{P}
    \fci[rectangle]{D}
    \end{influence-diagram}
         \caption{CDT}
         \label{fig:Newcomb-mech CDT}
     \end{subfigure}
     \hfill
     \begin{subfigure}[b]{0.3\linewidth}
         \centering
         \begin{influence-diagram}
    \node (D) [decision] {$D$};
    \node (P) [right = of D] {$P$};
    \node (U) [below right = 0.5 and 0.2 of D, utility] {$U$};
    \node (Dt) [relevanceb, above = 0.5 of D] {$\tilde{D}$};
    \node (Pt) [relevanceb, above = 0.5 of P] {$\tilde{P}$};

    \edge {D} {U};
    \edge {Dt} {D};
    \edge {Pt} {P};
     \edge {Dt} {Pt};
    \edge {P} {U};
     \edge[information] {P} {D};
    \node at (Dt) [draw, rectangle, incentive, fci-color, minimum size=9mm] {};
    \end{influence-diagram}
         \caption{FDT}
         \label{fig:Newcomb-mech FDT}
     \end{subfigure}

        \caption{The operation performed by (a) an evidential decision theorist, (b) a causal decision theorist, and (c) a functional decision theorist on a model of the Transparent Newcomb problem.}
        \label{fig:transparent-newcomb-decision-theories}
        
\end{figure}

\subsection{Evidential Decision Theory}

Evidential decision theory \cite{ahmed_evidence_2014,jeffrey_logic_1990} says you should condition on your decision, and can be distilled into the following slogan:

\begin{quote}
    \textbf{Slogan}. Choose the action with the best news value.
\end{quote}

In Newcomb's problem, if you care about the evidence your actions provide, you should one-box. Though the contents of the opaque box are fixed, your beliefs about them aren't. If you one-box, there's a $99\%$ chance the opaque box contains the million; if you two-box there's a $99\%$ chance it doesn't.

Though EDT is not as immediately intuitively compelling as CDT, in some situations EDT agents appear to fare better: for example, EDT agents almost always leave Newcomb's problem as millionaires, while CDT agents almost never do.
This is sometimes called the "If you're so rational, why ain'cha rich?" argument \cite{lewis_why_1981}.

In Transparent Newcomb, on the other hand, your decision no longer provides any information about the contents of the second box, since you already know it. EDT therefore agrees with CDT that you should two-box. In the Twin Prisoner's Dillema, co-operation is evidence your twin will co-operate, and defection is evidence your twin will defect. Mutual co-operation is preferable to mutual defection, so EDT recommends that you co-operate.

\paragraph{Expected Utility Calculations} An evidential decision theorist computes the expected utility of taking each action $d\in dom(D)$ as $$\mathbb{E}\left[ U\mid D=d,\bm{Obs}_D=\bm{obs}_D \right].$$ 

\Cref{fig: Newcomb op EDT} shows the operation performed by EDT in a graphical model of Newcomb's problem. Since there are no observed variables, we just condition on $D$. There is an active path from $D$ to $U$ via $P$ -- the agent's decision gives information about their decision rule, which in turn gives information about the prediction mechanism, the prediction, and finally the agent's utility. An EDT agent opts to one-box in order to maximise its expected utility via this backdoor path.

\Cref{fig:Newcomb-mech EDT} shows the operation performed by EDT in the case of Transparent Newcomb. Since $P$ is now observed, we condition on it, blocking the backdoor path from $D$ to $U$. The EDT agent two-boxes in order to maximise its expected utility via the direct link from $D$ to $U$.

The graphical models in \cref{fig:TwinPD CDT} and \cref{fig:TwinPD FDT} can both be used to apply EDT to the Twin Prisoner's Dilemma. In either case, conditioning on $D$ affects the value of $T$ via a backdoor path, and so the agent chooses to co-operate so that its twin does the same.

\paragraph{Model Requirements} Since EDT agents' expected utility calculations only involve associational queries, they just require a model at the first rung of Pearl's causal hierarchy. They do not require a mechanised model, but they can use one.

A model of a decision problem sufficient to apply EDT can be obtained more easily than in the case of CDT, since we only need to collect observational rather than interventional data. We can simply observe similar agents facing the same decision problem, record their observations, their decision, and the utility they recieve, and then estimate $P(U\mid D,\bm{Obs_D})$.

\subsection{Functional Decision Theory}

Functional decision theory, as presented in \cite{yudkowsky_functional_2018} and \cite{levinstein_cheating_2020}, can be summed up as follows:
\begin{quote}
\textbf{Slogan}. Choose the way you would like all instances of your decision function to behave.    
\end{quote}

We propose to think of FDT in a slightly different way, as a version of CDT with two key differences: firstly, FDT chooses the best \textit{decision rule} to follow, rather then the best action to take, and secondly, FDT uses a logical rather than physical notion of causality. 
Using this idea, we can suggest a new definition of FDT in mechanised causal graphs.

\paragraph{Expected Utility Calculations} A functional decision theorist computes the expected utility of taking each action $d\in dom(D)$ as $$\mathbb{E}\left[ U\mid do(\tilde{D}=\tilde{d}) \right].$$

In Newcomb's problem (\cref{fig: Newcomb op FDT}), intervening on $\tilde{D}$ affects $U$ via both $D$ and $P$. In Transparent Newcomb (\cref{fig:Newcomb-mech FDT}), the fact that FDT does not condition on observed variables means the path to $U$ via $P$ remains unblocked, and so an FDT agent one-boxes in both. 

The property of not conditioning on observed variables is known as "updatelessness" (Section \ref{sec:discussion}), and it is this property, rather than FDT's alternative notion of causality, which explains its behaviour here. An updateless version of CDT would one-box as well, since the physical and logical-casual models are the same.

But the logical view of causality \textit{is} relevant in the Twin Prisoner's Dilemma. An FDT agent treats their twin's policy as a logical consequence of their own, so that the problem becomes analogous to Newcomb's problem (see \cref{fig:TwinPD FDT}). In particular, this means that the twin's decision about whether to co-operate is downstream of the variable on which FDT intervenes, and so an FDT agent follows a policy of co-operation in order to influence the twin. Note that policy-selection alone is not enough to achieve co-operation here, since in a physical causal model the twin's policy is \textit{not} downstream of the agent's own.

\paragraph{Model Requirements.} FDT agents require a model at the second rung of Pearl's causal hierarchy, and they need it to involve their decision rule variable $\tilde{D}$. 

However, in contrast to CDT, FDT agents require a model of the interventional distributions under interventions on \textit{logical} rather than physical variables. While we can often intuitively imagine how the world would look under a logical intervention -- if all existing instances of an agent's source code compiled in different way, for example -- the notion is not sufficiently formalised that we can say how one could actually go out into the world and determine such interventional distributions from observation and experiment. In other words, FDT has an epistemic problem -- we cannot yet say how FDT agents can find out what they need to know, even in principle. 
For this, something like a formal theory of logical causality would be needed.

\subsection{A Discussion of Updatelessness}
To motivate FDT, consider that the "Why ain'cha rich?" argument can be turned on both EDT and CDT in cases where agents would benefit from being able to make binding precomitments, such as Transparent Newcomb.

EDT and CDT agents alike reason that since you can see that the second box is full, you are choosing between a sure million and a sure one million one thousand. However, the second box will only \textit{be} full if the predictor happens to have predicted that you would one-box. Since CDT and EDT agents in fact two-box, and since the predictor is right 
$99\%$ of the time, they will almost always be choosing between a sure nothing and a sure one thousand. We can calculate the expected winnings of a CDT or EDT agent facing Newcomb's problem as 
\$$11\ 000$\footnote{$99\%\times$\$$1\ 000+1\%\times $\$$1\  000\ 000$.}. 
Given the opportunity to make a binding precommitment to one-boxing if the second box is full, either type of agent would take it, since doing so would up their expected winnings to 
\$$990\ 010$
\footnote{$99\%
\times\text{\$}
1\ 000\ 000+1\%\times
\text{\$}1000$
(the agent is free to take the thousand if the million is not present).}.

It is insightful to notice that an agent that \textit{acts as it would ideally have precommitted to act} can achieve exactly the same outcomes even in the absence of the opportunity to precommit. Suppose an agent finds itself in the Transparent Newcomb problem, not having precommitted to anything, realises it would have liked to have precommited to one-boxing upon seeing the million, and decides to act in accordance with that commitment anyway. With $99\%$ probability the predictor will have predicted this and the agent will come away a millionaire.

Several proposals formalise this idea in terms of selecting the best policy to follow, rather than the best action to take \cite{gauthier_assure_1994,meacham_binding_2010},  or in terms of not performing a Bayesian update on your beliefs after making an observation \cite{dai_towards_2009}. The proposals amount to almost the same idea, and in the framework of mechanised causal graphs it becomes clear that in order to act as you would ideally have precommitted to act it is necessary to \textit{both} switch from selecting your decision $D$ to your decision rule $\tilde{D}$, \textit{and} stop conditioning on your observations $\bm{Obs}_D$.

We prefer to call this idea "updatelessness" \cite{dai_towards_2009} rather than "policy-selection", primarily because selecting the policy while conditioning on observations will not have the desired effect. Indeed, it is somewhat difficult to construct a reasonable decision problem in which conditioning on $\bm{Obs_D}$ while selecting $\tilde{D}$ is different from conditioning $\bm{Obs_D}$ while selecting $D$. It might therefore be most natural to think of all of the decision theories discussed here as ways of calculating of the expected utility of following each decision \textit{rule}, with some of them conditioning these calculations on the current observation, and some not. However, in order not to stray to far from ordinary convention, we will continue to notate updateful decision theories as selecting $D$.

Mechanised causal graphs clarify that the original definition of functional decision theory (as choosing the output of all instances of your decision function) can be thought of as a combination of updatelessness with an alternative notion of causality\footnote{FDT as presented in \cite{yudkowsky_functional_2018} selects the action the decision rule outputs on the current observation, rather than the whole decision rule. This should not be confused with updatefulness. For simplicity we treat it as selecting the whole decision rule.}.

A final point to note is that issues of updatelessness can only arise in models of decision-making where agents are asked to choose \textit{conditional} policies, such as in extensive-form games, or the graphical models we consider here. When we represent a game in normal-form, we effectively stipulate that there is a single decision to be made by each player, unconditional on any observations. As in the Twin Prisoner's Dilemma, updatelessness will not affect an agent's decision in a "twin" version of \textit{any} normal-form game.

\section{A Taxonomy of Decision Theories}
\label{taxonomy}

In this section we categorise the decision theories considered so far along two axes, and briefly discuss the decision theories occupying the other positions.
\paragraph{The Evidential/Causal/Functional Axis}
A decision theory is
\begin{itemize}
    \item \emph{evidential} if it computes expected utility by conditioning: $$\mathbb{E}\left[U\mid\dots\right].$$
    \item \emph{causal} if it computes expected utility by intervening in a causal model: $$\mathbb{E}\left[U\mid do(\dots)\dots\right].$$
    \item \emph{functional} if it computes expected utility by intervening in a logical-causal model: $$\mathbb{E}\left[U\mid do(\dots)\dots\right].$$
\end{itemize}

\paragraph{The Updateless/Updateful Axis}

A decision theory is
\begin{itemize}
    \item \emph{updateful} if it selects $D$ and conditions on $\bm{Obs}_D$:
    $$\mathbb{E} \left[ U \mid \dots D \dots \textbf{\emph{Obs}}_D \right].$$
    
    \item \emph{updateless} if it selects $\tilde{D}$ (without conditioning on $\bm{Obs_D}$):  $$\mathbb{E}\left[U\mid\dots\tilde{D}\dots\right].$$
\end{itemize}

Altogether there are six positions on these axes, each of which corresponds to a distinct decision theory. \Cref{fig: EU table.} summarises their respective expected utility calculations, and \cref{fig: Behaviour table.} shows how each behaves in the decision problems we've considered.
\begin{table}
\centering
\footnotesize
\begin{tabular}{|l|l|l|l|}
\hline
 & Evidential & Causal & Functional \\ \hline
Updateful & $\mathbb{E}\left[U\mid D, \bm{Obs}_D\right]$ & \begin{tabular}[c]{@{}l@{}}$\mathbb{E}\left[U\mid do(D),\bm{Obs}_D\right]$\\ in a causal model.\end{tabular} & \begin{tabular}[c]{@{}l@{}}$\mathbb{E}\left[U\mid do(D),\bm{Obs}_D\right]$\\ in a logical-causal model.\end{tabular} \\ \hline
Updateless &  $\mathbb{E}\left[U\mid \tilde{D}, \right]$ & \begin{tabular}[c]{@{}l@{}}$\mathbb{E}\left[U\mid do(\tilde{D})\right]$\\ in a causal model.\end{tabular} & \begin{tabular}[c]{@{}l@{}}$\mathbb{E}\left[U\mid do(\tilde{D})\right]$\\ in a logical-causal model.\end{tabular} \\ \hline
\end{tabular}
\caption{The expected utility calculations performed by each decision theory.}
\label{fig: EU table.}
\end{table}
\begin{table}
\centering
\footnotesize
\begin{tabular}{|l|l|l|l|l|}
\hline
 & Newcomb & Transparent Newcomb & Twin PD  \\ \hline
EDT & one-box & two-box & co-operate  \\ \hline
CDT & two-box & two-box & defect  \\ \hline
Updateful FDT & one-box & two-box & co-operate  \\ \hline
Updateless EDT & one-box & one-box & co-operate  \\ \hline
Updateless CDT & one-box & one-box & defect  \\ \hline
FDT & one-box & one-box & co-operate  \\ \hline
\end{tabular}
\caption{The behaviour of the each decision theory in the example decision problems.}
\label{fig: Behaviour table.}
\end{table}

The axes give us a natural naming scheme: "(updateful|updateless) (evidential|causal|functional) decision theory". Unfortunately our existing terminology might confuse things: "EDT" and "CDT" can be thought of as shortenings of "updateful EDT" and "updateful CDT" respectively, while "FDT" should be thought of as a shortening of "\textit{updateless} FDT". We will abbreviate "updateless EDT" as  "UEDT" and "updateless CDT" as "UCDT", but to avoid confusion we will not abbreviate "updateful FDT" any further.

We will briefly mention a few key points about the other decision theories in the table.

\textbf{Updateless CDT}.
Updateless CDT has been frequently discussed in the literature under names such as policy-CDT \cite{oesterheld_extracting_2021}. It requires a causal model over a set of variables that includes the agent's decision rule. Such a model can be obtained by following the experimental procedure for obtaining an unmechanised model, but randomly assigning decision rules, rather than actions, to observed agents.

\textbf{Updateless EDT}.
Since it only requires an associational model over a set of variables that includes the decision rule, UEDT shares none of FDT's epistemic issues, but only seems to diverge from its decision-making on problems which philosophers consider questionably well-defined, such as the smoking lesion problem \cite{oesterheld_understanding_2018}. For that reason UEDT is probably an adequate substitute for FDT, both for the purposes of decision-making and for analysing the incentives of ML systems.

\textbf{Updateful FDT}.
To our knowledge, an updateful variant of FDT has not previously appeared in the literature, although it bears some similarity to \cite{spohn_reversing_2012}. It shares the epistemic problems of its updateless cousin. Intuitively we can transform a model suitable for FDT into a model suitable for updateful FDT by reversing the arrow from $\tilde{D}$ to $D$ -- choosing your decision should logically affect your decision rule. In contrast to the updateless case, the updateful variants of FDT and EDT intuitively disagree on uncontroversial problems such as XOR Blackmail \cite{soares_toward_2015,levinstein_cheating_2020}.

\subsection{Comparison To Garrabrant's Axes}
Our axes are essentially a simplified version of the axes informally discussed  
in \cite{garrabrant_miriop_2021}, which make three distinctions: a decision theory can be evidential or causal, updateful or updateless, and can model the world using a \textit{physical} or \textit{algorithmic} agent ontology. The final axis corresponds to the question of whether you imagine you are controlling a logical variable (the output of an algorithm) or a physical one. In other words, whether you are choosing for \textit{you}, \textit{right here right now}, or whether you are choosing for \textit{agents like you} in \textit{situations like this}.

There are eight possible positions on these axes, but it is not clear that each position corresponds to a distinct decision theory. In particular, the physical vs algorithmic agent ontology axis appears irrelevant to evidentialists: evidence about what you decide, right here right now, is evidence about what agents like you decide in situations like this and vice versa.\footnote{Wei Dai's \textit{updateless decision theory} \cite{dai_towards_2009}, is intended to be evidential and updateless with an algorithmic ontology, and seemingly disagrees with UEDT on problems like the Smoking Lesion. Since UDT and FDT are not formally defined it is hard to compare the two, but their intended notions of dependence appear to coincide, and we prefer to think of it as logical-causal rather than evidential.}

We therefore prefer to collapse the third axis into the first and think of the ontological question as being about two competing notions of causality. We do not claim that our axes are more fundamentally meaningful than Garrabrant's, but they allow for a simplified presentation which removes some redundancy, and a naming system which lines up well with existing terminology.

\section{Well-Defining Transparent Newcomblike Problems}
\label{defining}
\label{sec:discussion}
Decision problems that involve an agent being shown the result of a prediction about their action before choosing it are sometimes considered to be questionably well-defined, since it's possible for an agent to choose in such a way that falsifies the prediction. For example, in some versions of Transparent Newcomb, the predictor is said to put the million in the box if and only if the it predicts the agent will one-box \cite{arntzenius_no_2008,gauthier_xiineighbourhood_1989}. But if the agent follows the decision rule "one-box if and only if the million is not present", the prediction will \textit{always} be wrong, and the premise that the predictor is highly accurate is falsified.

The author of \cite{meacham_binding_2010} gets around this by stipulating that agents are banned from following such policies. An alternative approach \cite{drescher_good_2006,yudkowsky_functional_2018}, and the one used in this paper, is to have the predictor predict what the agent will choose \textit{upon receiving a certain observation}, rather than what the agent will do full stop. In this version of Transparent Newcomb, the predictor fills the box if and only if it predicts the agent will one-box \textit{upon seeing the box is full}.

Using mechanism variables, we can adopt a general principle which clarifies why this works and ensures we never run into the issue when defining decision problems. We say that accurate predictions should be thought of as depending causally, logically or evidentially on an agent's \textit{decision rule} rather than on their decision. Predicting how an agent will respond to one specific observation is a special case of making a prediction about their decision rule.
\balance 
Intuitively this avoids the circular situation where the decision causes the prediction and the prediction causes the decision. But note that the issue is not fundamentally about causality: even the associational proposition that in $99\%$ of cases where the agent one-boxes, the predictor fills the box, and in $99\%$ of cases where the agent two-boxes, the box is left empty, is falsified by the agent following the aforementioned policy.

This becomes especially clear when we try to represent the situation graphically. The two conditional distributions cannot represented with a DAG -- we inevitably get a cycle, as in \cref{fig: cyclic TN}. On the other hand, we can posit any conditional distribution of the prediction given the decision \textit{rule}, alongside any conditional distribution of the decision given the prediction, without getting a cycle.

\section{Conclusions}

In this paper, we have shown how mechanised causal graphs can be used to characterise a range of different decision theories, and the axes along which they differ.
We hope this analysis will be a step towards a clearer understanding of machine learning systems, and their interactions with other agents and humans.

In particular, this paper may enable incentive analysis of agents with different decision theories.
Previous work has analysed the incentive structures of agents using causal decision theory in different machine learning setups \cite{everitt_agent_2021}. But agents following different decision theories will face different incentives. Mechanised causal graphs should allow us to combine the two ideas and analyse the incentives structures in each case.

\subsubsection{Acknowledgements}
 The authors would like to thank Caspar Oesterheld, Ryan Carey, Scott Garrabrant, and Francis Rhys Ward for feedback and discussion.
 This work was supported by UK Research and Innovation [grant number EP/S023356/1], in the UKRI Centre
 for Doctoral Training in Safe and Trusted Artificial Intelligence.

\bibliographystyle{splncs04}

\bibliography{references,references-tom}
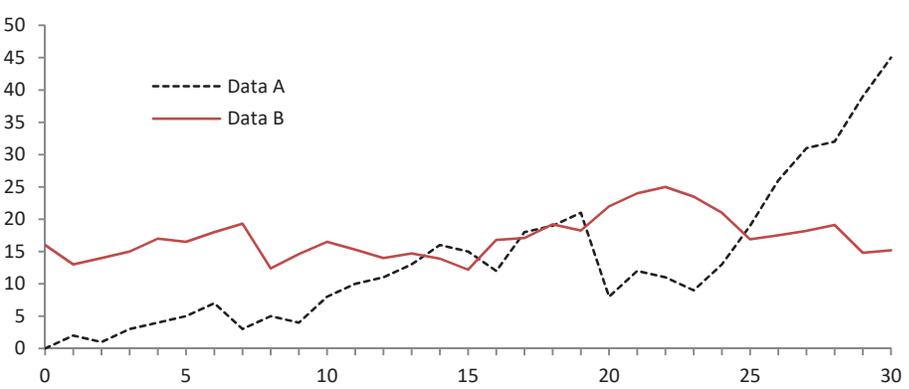
\end{document}